# Distribution-Specific Agnostic Boosting

Vitaly Feldman

*IBM Almaden Research Center*

**Abstract**

We consider the problem of boosting the accuracy of weak learning algorithms in the agnostic learning framework of Haussler (1992) and Kearns *et al.* (1992). Known algorithms for this problem (Ben-David et al., 2001; Gavinsky, 2002; Kalai *et al.*, 2008) follow the same strategy as boosting algorithms in the PAC model: the weak learner is executed on the same target function but over different distributions on the domain. Application of such boosting algorithms usually requires a distribution-independent weak agnostic learners. Here we demonstrate boosting algorithms for the agnostic learning framework that only modify the distribution on the labels of the points (or, equivalently, modify the target function). This allows boosting a distribution-specific weak agnostic learner to a strong agnostic learner with respect to the same distribution. Our algorithm achieves the same guarantees on the final error as the boosting algorithms of Kalai *et al.* (2008) but is substantially simpler and more efficient.

When applied to the weak agnostic parity learning algorithm of Goldreich and Levin (1989) our algorithm yields a simple PAC learning algorithm for DNF and an agnostic learning algorithm for decision trees over the uniform distribution using membership queries. These results substantially simplify Jackson's famous DNF learning algorithm (1994) and the recent result of Gopalan *et al.* (2008).

We also strengthen the connection to hard-core set constructions discovered by Klivans and Servedio (1999) by demonstrating that hard-core set constructions that achieve the optimal hard-core set size (given by Holenstein (2005) and Barak *et al.* (2009)) imply distribution-specific agnostic boosting algorithms. Conversely, our boosting algorithm gives a simple hard-core set construction with an (almost) optimal hard-core set size.

## 1 Introduction

A boosting algorithm is a technique for combining the outputs of a learning algorithm(s) of low but non-trivial accuracy to obtain a hypothesis of high(er) accuracy. Since its introduction by Schapire [31] in the Valiant's PAC learning model [33] it has become one of most studied areas in the theoretical and applied machine learning and also one of the tools widely used in practice.

While numerous boosting algorithms are known that can boost the accuracy of a weak PAC learner [26] to an arbitrarily high value, very few boosting algorithms can provably improve the accuracy in the presence of noisy or inconsistent data[1].

A natural model of learning without the PAC model assumptions on the target function is the agnostic learning model of Haussler [16] and Kearns, Schapire and Sellie [25]. The goal of an agnostic learning algorithm for a concept class $C$ is to produce, for any distribution on examples, a hypothesis $h$ whose error on the distribution is close to the best possible by a concept from $C$. This model reflects a common empirical approach to learning, where few or no assumptions are made on the process that generates the examples and a limited space of candidate hypothesis functions is searched in an attempt to find the best approximation to the given data.

---

[1] A number of boosting algorithms were designed specifically to address the suboptimal performance of early boosting algorithms on noisy data. However they were either still analyzed in the noiseless PAC model (e.g. [10]) or only tested empirically.

The problem of boosting the accuracy of a weak learner in the agnostic learning framework was first considered by Ben-David, Long and Mansour [2]. The weak learner that was given to the boosting algorithm in their definition was a $\beta$-optimal agnostic learner, namely an agnostic learner that for any distribution $A$, produces a hypothesis with error $\Delta + \beta$, where $\Delta$ is the error of the best hypothesis in $C$ (relative to $A$). Ben-David *et al.* described a boosting algorithm that for a certain range of values $\Delta$ and $\beta$ produces a hypothesis that has a lower error than the provided weak learner. In a subsequent work Gavinsky showed that a $\beta$-optimal agnostic learner can be boosted to a learner that achieves the error of $\frac{\Delta}{1/2-\beta} + \epsilon$ in time polynomial in $1/\epsilon$ [12]. He has also shown that this error is within the factor of 2 from the best achievable for this problem.

Recently Kalai, Mansour and Verbin have examined boosting a different type of weak learner [22]. Specifically, they define an $(\alpha, \gamma)$-weak agnostic learner to be a learning algorithm that produces a hypothesis with the error of at most $1/2 - \gamma$ whenever $\Delta \leq 1/2 - \alpha$. Kalai *et al.* gave a boosting algorithm that boosts any $(\alpha, \gamma)$-weak agnostic learner to an $(\alpha + \epsilon)$-optimal agnostic learner in time polynomial in $1/\gamma$ and $1/\epsilon$. They have also demonstrated that such a boosting algorithm can be used to obtain the first non-trivial distribution-independent agnostic learning algorithm for parities. Their boosting algorithm is based on a boosting-by-branching-programs algorithm of Mansour and McAllester [30] and its analysis by Kalai and Servedio [23].

As these agnostic boosting algorithms are based on boosting algorithm in the PAC learning framework, they work by applying the weak learner to the target function on carefully constructed distributions over the domain. This implies that such boosting algorithms can only be applied in the distribution-independent setting. (One notable exception to this rule is Jackson's algorithm for learning DNF over the uniform distribution [19] that boosts the accuracy via an ad hoc extension of the weak learner of Blum *et al.* [3] to distributions that are close to the uniform).

## 1.1 Our Results

We present a simple *distribution-specific* agnostic boosting algorithm for $(\alpha, \gamma)$-weak agnostic learners. That is, our boosting algorithm does not modify the marginal distribution over the domain of the learning problem but instead modifies the distribution on the label of each example.

**Theorem 1.1** *There exists an algorithm* `ABoost` *that for every concept class $C$ and distribution $D$ over $X$, given an $(\alpha, \gamma)$-weak agnostic learning algorithm $\mathcal{A}$ for $C$ over $D$, agnostically and $\alpha$-optimally learns $C$ over $D$. Further,* `ABoost` *invokes $\mathcal{A}$ $O(\gamma^{-2})$ times and runs in time $poly(T, 1/\gamma, 1/\epsilon)$, where $T$ is the running time of $\mathcal{A}$.*

Our boosting algorithm implies that weak agnostic learning with respect to any specific distribution is equivalent to (strong) agnostic learning with respect to the same distribution (see Theorem 3.2 for the formal statement). An immediate application of this result is a simple agnostic learning algorithm for decision lists over the uniform distribution using membership queries (see Lemma 3.3). Recently, Gopalan, Kalai and Klivans gave the first algorithm for this problem [14]. Their proof is based on a substantially more involved and delicate argument.

In Section 3.2 we use our boosting algorithm to extend the observation that agnostic learning of a class $C$ implies PAC learning of low-weight linear thresholds of functions from $C$ [25, 9, 28] to a distribution specific setting. For a set of functions $C$ and integer $W$ denote by $\text{TH}(W, C)$ the set of all functions representable as $\text{sign}(\sum_{i \leq W} f_i(x))$ where for all $i$, $f_i \in C$.

**Theorem 1.2** *If $C$ is efficiently agnostically learnable with respect to distribution $D$ then $\text{TH}(W, C)$ is efficiently PAC learnable over $D$ for any $W$ upper-bounded by a polynomial in the learning parameters.*

An immediate application of this result is a simple proof that DNF are learnable over the uniform distribution using membership queries [19] (we include the details in Section 3.2). It also allows to simplify the analysis in many subsequent algorithms for learning DNF that use the same boosting-based approach (e.g. [4, 5, 9]). In addition, this result gives a new implication of an agnostic algorithm for learning DNF that is posed as an open problem by Gopalan *et al.* [14].

We show that our boosting algorithm can also be viewed in a more traditional setting where the boosting algorithm runs the weak learner on modified marginal distribution but does not modify the label distribution. In particular, in the setting of Ben-David *et al.* [2] our boosting algorithm achieves the optimal accuracy of $\frac{\Delta}{1-2\beta} + \epsilon$ (or 1/2 of the accuracy achieved by Gavinsky's boosting algorithm [12]). The details of this version are given in Section 3.1.

Boosting algorithms are also known to be closely related to hard-core set constructions [27], a technique in hardness amplification [18]. Given a function $f$ that cannot be $\tau$-approximated on $X$ by circuits of certain size $s$ the goal of a hard-core set construction is to construct a sufficiently large subset of $X$ on which $f$ cannot be $(1/2-\gamma)$-approximated by circuits of a slightly smaller than $s$ size. Here we strengthen the connection discovered by Klivans and Servedio [27] by observing that hard-core set constructions achieving the optimal hard-core set size of $2\tau$ give agnostic boosting algorithms. The first construction with this property was given by Holenstein who used the construction to obtain a key agreement protocol from a weak bit agreement primitive [17]. In a recent work Barak, Hardt and Kale demonstrated a more efficient hard-core construction with this property [1]. Both of these constructions can be easily translated into agnostic boosting algorithms. In addition, we show that our agnostic boosting algorithm gives a new hardcore set construction algorithm with an almost the optimal hard-core set size parameter. Our technique of achieving the optimal hard-core set size is different from the method of Holenstein [17] (which is also used by Barak *et al.* [1]) and the resulting algorithm is simpler to analyze. The relation to hard-core set constructions is presented in Section 4.

## 1.2 Techniques

Our boosting algorithms build a hypothesis $h : X \to [-1, 1]$ in steps starting from the $h_0 \equiv 0$ hypothesis. At step $i$ the weak learner is run on points drawn randomly from the base distribution $D$ and the labels given by $(f(x) - h_i(x))/2$, that is the expectation of the random $\{-1, 1\}$ label assigned to point $x$ is $(f(x) - h_i(x))/2$. A weak hypothesis for this distribution satisfies $\mathbf{E}_D[(f(x) - h_i(x)/2)g(x)] \geq 2\gamma$. We let $h'_{i+1} = h_i + \gamma \cdot g$. It is easy to see, that after this step $h'_{i+1}$ is "closer" to $f$ than $h_i$ when the functions are viewed as vectors in the appropriate Euclidean space. This argument requires the hypothesis at each step to have range in $[-1, 1]$ and therefore we apply a projection step. Namely $h_{i+1}$ is obtained from $h'_{i+1}$ by cutting off all values outside the range $[-1, 1]$. This step only reduces the distance. This algorithm and the distribution-specific view of boosting are implicit in [8] where the algorithm is used to characterize the query complexity of statistical query (SQ) [24] learning (in the PAC and the agnostic models) using a characterization of weak SQ learning. However the algorithm we described so far can only guarantee a hypothesis with the error equal to twice the optimum. To achieve the optimum we add new "balancing" steps to process. Namely, we test the hypothesis $-\text{sign}(h_i)$ on the data distribution produced at step $i$. If this hypothesis has non-trivial performance it is used to update $h_i$ in the same way as the weak learner. Otherwise, it is easy to show that $\text{sign}(h_i)$ has half the error of $h_i$ itself which is close to the optimum at the end of the boosting process.

For our application to distribution-independent boosting and hard-core set construction we also give a stronger boosting algorithm that uses the same argument but on the basis of a slightly different way to produce distribution together with a corresponding distance function (see Theorem 1.1).

## 1.3 Related Work

Kalai and Kanade have very recently and independently demonstrated a different distribution-specific agnostic boosting algorithm [20]. Their boosting algorithm is based on a smooth version of Adaboost [11] by Domingo and Watanabe [7] and Servedio [32] and uses an equivalent of our "balancing" step. It requires a similar number of boosting stages and running time as our `ABoost` algorithm. They also show an analogous application to agnostic learning of decision trees (see Lemma 3.3). In addition Kalai and Kanade give a simpler version of the agnostic halfspace learning algorithm of Kalai *et al.* [21] and include results from an empirical evaluation of their algorithm.

## 2 Preliminaries

Let $X$ denote some fixed domain and let $\mathcal{F}_1^\infty$ denote the set of all functions from $X$ to $[-1, 1]$ (that is all the functions with $L_\infty$ norm bounded by 1). It will be convenient to view a distribution $D$ over $X$ as defining the product $\langle \phi, \psi \rangle_D = \mathbf{E}_{x \sim D}[\phi(x) \cdot \psi(x)]$ over the space of real-valued functions on $X$. It is easy to see that this is simply a non-negatively weighted version of the standard dot product over $\mathbb{R}^X$ and hence is a positive semi-inner product over $\mathbb{R}^X$. The corresponding norm is defined as $\|\phi\|_D = \sqrt{\mathbf{E}_D[\phi^2(x)]} = \sqrt{\langle \phi, \phi \rangle_D}$.

### 2.1 Agnostic Learning

The *agnostic* learning model was introduced by Haussler [16] and Kearns *et al.* [25] in order to model situations in which the assumption that examples are labeled by some $f \in \mathcal{C}$ does not hold. In its least restricted version the examples are generated from some unknown distribution $A$ over $X \times \{-1, 1\}$. The goal of an agnostic learning algorithm for a concept class $C$ is to produce a hypothesis whose error on examples generated from $A$ is close to the best possible by a concept from $C$. Any distribution $A$ over $X \times \{-1, 1\}$ can be described uniquely by its marginal distribution $D$ over $X$ and the expectation of $b$ given $x$. That is, we refer to a distribution $A$ over $X \times \{-1, 1\}$ by a pair $(D_A, \phi_A)$ where $D_A(z) = \mathbf{Pr}_{\langle x, b \rangle \sim A}[x = z]$ and
$$\phi_A(z) = \mathbf{E}_{\langle x, b \rangle \sim A}[b \mid z = x].$$

Formally, for a Boolean function $h$ and a distribution $A = (D, \phi)$ over $X \times \{-1, 1\}$, we define
$$\Delta(A, h) = \mathbf{Pr}_{\langle x, b \rangle \sim A}[h(x) \neq b].$$

We will frequently use the following simple equality $\Delta(A, h) = (1 - \langle \phi, h \rangle_D)/2$. For a concept class $C$, define $\Delta(A, C) = \inf_{h \in C} \{\Delta(A, h)\}$.

Kearns *et al.* [25] define agnostic learning as follows.

**Definition 2.1** *An algorithm $\mathcal{A}$ agnostically learns a concept class $C$ by a representation class $H$ if for every $\epsilon > 0, \delta > 0$, distribution $A$ over $X \times \{-1, 1\}$, $\mathcal{A}$ given access to examples drawn randomly from $A$, outputs, with probability at least $1 - \delta$, a hypothesis $h \in H$ such that $\Delta(A, h) \leq \Delta(A, C) + \epsilon$.*

As in the PAC learning, the learning algorithm is *efficient* if it runs in time polynomial $1/\epsilon, \log(1/\delta)$ and $n$. Here and elsewhere when not noted otherwise, we use $n$ as a bound on the description length of every concept in $C$ and also the dimension of the domain.

In the distribution-specific version of this model, learning is only required for every $A = (D, \phi)$, where $D$ equals to some fixed distribution known in advance.

In order to define boosting in the agnostic setting we use the following definitions from [22]. For $0 < \beta \leq 1/2$ we say that a learning algorithm is $\beta$-optimal agnostic if for every $\epsilon > 0$ the algorithm produces a hypothesis $h$ such that $\Delta(A, h) \leq \Delta(A, C) + \beta + \epsilon$. We note that Ben-David *et al.* [2] use a slightly stronger notion of $\beta$-optimality that does not have the extra $\epsilon$ but this will not be significant for our discussion.

For $0 < \gamma \leq \alpha \leq 1/2$ we say that a learning algorithm is $(\alpha, \gamma)$-weak agnostic if the algorithm produces a hypothesis $h$ such that $\Delta(A, h) \leq 1/2 - \gamma$ whenever $\Delta(A, C) \leq 1/2 - \alpha$.

For convenience when discussing weak agnostic learning we also define $\Gamma(A, h) = 1/2 - \Delta(A, h)$ and $\Gamma(A, h) = 1/2 - \Delta(A, C)$ accordingly. A *weak agnostic* learning algorithm is an algorithm that can recover at least a polynomial fraction of the advantage over the random guessing of the best approximating function in $C$. Specifically, it produces a hypothesis $h$ such that $\Gamma(A, h) \geq p(1/n, \Gamma(A, C))$ for some polynomial $p(\cdot, \cdot)$.

## 3 Agnostic Boosting

The main component of the agnostic boosting algorithm in the work of Kalai *et al.* [22] is a more general algorithm that boosts every $(\alpha, \gamma)$-weak agnostic learner to an $\alpha$-optimal agnostic learner. We first show a weaker algorithm that boosts any $(\alpha,\gamma)$-weak agnostic learner to a $2\alpha$-optimal agnostic learner (but is sufficient for boosting a weak agnostic learning algorithm to a strong one).

**Theorem 3.1** *There exists an algorithm* `A2boost` *that for every concept class $C$ and distribution $D$ over $X$, given an $(\alpha,\gamma)$-weak agnostic learning algorithm $\mathcal{A}$ for $C$ over $D$ agnostically and $2\alpha$-optimally learns $C$ over $D$. Further,* `A2boost` *invokes $\mathcal{A}$ $O(\gamma^{-2})$ times and runs in time $\mathrm{poly}(T, 1/\gamma, 1/\epsilon)$, where $T$ is the running time of $\mathcal{A}$.*

**Proof:** Let $A = (D, \phi)$ be the target distribution over examples. Our algorithm performs a form of gradient descent to the unknown target function $\phi$, where the weak agnostic learning provides the equivalent of gradient computation.

We start with a hypothesis $h_0 \equiv 0$. Let $h_i \in \mathcal{F}_1^\infty$ be the current hypothesis. We run the algorithm $\mathcal{A}$ on examples from $A_i = (D, (\phi - h_i)/2)$. Note that $(\phi - h_i)/2 \in \mathcal{F}_1^\infty$ and therefore this is possible. If a hypothesis $g$ with error of at most $1/2 - \gamma$ is output by $\mathcal{A}$ we update $h_i$ using $g$ in the way we describe later. Otherwise, we test the error of $-\mathrm{sign}(h_i)$ on distribution $A_i$. If the error is at most $1/2 - \epsilon/2$ then we update $h_i$ using $-\mathrm{sign}(h_i)$. We refer to this update as *balancing*. If neither of these conditions holds the algorithm stops and outputs $\mathrm{sign}(h_i)$ as its final hypothesis. To reduce the number of potentially more expensive invocations of the weak learner we perform balancing steps until $\Delta(A_i, -\mathrm{sign}(h_i)) \geq 1/2 - \epsilon/2$.

To update $h_i$ using a function $g_i$ (which is either $g$ or $-\mathrm{sign}(h_i)$) that has error $1/2 - \gamma_i$ we add $\gamma_i \cdot g_i$ to $h_i$ and then truncate all the values outside of $[-1,1]$. Namely we set $h'_{i+1} = h_i + \gamma_i \cdot g_i$ and let $h_{i+1} \equiv P_1(h'_{i+1})$, where

$$P_1(a) \triangleq \begin{cases} a & |a| \leq 1 \\ \mathrm{sign}(a) & \text{otherwise.} \end{cases}$$

We note that when $g_i = -\mathrm{sign}(h_i)$ the projection step $P_1$ will not be necessary since $h'_{i+1} \in \mathcal{F}_1^\infty$.

We first prove that this process will terminate after at most $O(\gamma^{-2})$ invocations of the weak learner and $O(\epsilon^{-2})$ balancing steps. To show this we prove that in each step $h_i$ is closer to $\phi$ by at least $3\gamma_i^2$. Specifically, we claim that

$$\|\phi - h_{i+1}\|_D^2 \leq \|\phi - h_i\|_D^2 - 3\gamma_i^2 \ .$$

By the definition, $g_i$ has error $1/2 - \gamma_i$ on $A_i$. This is equivalent to $\langle \phi - h_i, g_i \rangle_D \geq 2\gamma_i$. Therefore

$$\|\phi - h'_{i+1}\|_D^2 = \|\phi - (h_i + \gamma_i g_i)\|_D^2 = \|\phi - h_i\|_D^2 - 2\gamma_i \langle \phi - h_i, g_i \rangle_D + \gamma_i^2 \|g_i\|_D^2 \leq \|\phi - h_i\|_D^2 - 4\gamma_i^2 + \gamma_i^2$$
$$= \|\phi - h_i\|_D^2 - 3\gamma_i^2$$

We now observe that the projection step can only decrease the distance to $\phi$, in other words $\|\phi - P_1(h'_{i+1})\|_D^2 \leq \|\phi - h'_{i+1}\|_D^2$. This follows easily from the fact that for any value $b \in [-1,1]$ and any real value $a$, $(b - P_1(a))^2 \leq (b-a)^2$. Hence, $\|\phi - h_{i+1}\|_D^2 \leq \|\phi - h'_{i+1}\|_D^2$.

By the definition, after each successful invocation of the weak learner $\gamma_i = \gamma$ and at most one not successful invocation of the weak learner is performed for every successful one. Similarly, in each balancing update $\gamma_i = \epsilon/2$ and at most two tests of the error of $-\mathrm{sign}(h_i)$ are performed for each balancing update. In addition, $\|\phi - h_0\|_D^2 = \|\phi\|_D^2 \leq 1$ and therefore the process has to terminate after at most $2\gamma^{-2}/3$ invocations of the weak learner and $4\epsilon^{-2}/3$ balancing updates.

We now need to prove that the final hypothesis $h = \mathrm{sign}(h_t)$ satisfies $\Delta(A,h) \leq \Delta(A,C) + 2\alpha + \epsilon$. Let $c \in C$ be the function such that $\Delta(A,C) = \Delta(A,c)$ or $\langle \phi, c \rangle_D = 1 - 2\Delta(A,C)$. By the definition of the final hypothesis $h$, we know that our boosting algorithm has not received a weak hypothesis with error $\leq 1/2 - \gamma$. By the property of $(\alpha,\gamma)$-weak agnostic learning this implies that $\Delta(A_t, C) \geq \Delta(A_t, c) \geq 1/2 - \alpha$, or $\langle (\phi - h_t)/2, c \rangle_D \leq 2\alpha$. This gives us that $\langle h_t, c \rangle_D \geq \langle \phi, c \rangle_D - 4\alpha = 1 - 2\Delta(A,C) - 4\alpha$ $(*)$.

In addition, we know that the error of $-\mathrm{sign}(h_t)$ on $A_t$ is at least $1/2 - \epsilon/2$. That is, $\langle (\phi - h_t)/2, -\mathrm{sign}(h_t) \rangle_D \leq \epsilon$. This gives $\langle \phi, \mathrm{sign}(h_t) \rangle_D \geq \langle h_t, \mathrm{sign}(h_t) \rangle_D - 2\epsilon$. We observe that $\langle h_t, \mathrm{sign}(h_t) \rangle_D \geq \langle h_t, c \rangle_D$ and combine this with $(*)$ to obtain

$$\langle \phi, h \rangle_D = \langle \phi, \mathrm{sign}(h_t) \rangle_D \geq \langle h_t, \mathrm{sign}(h_t) \rangle_D - 2\epsilon \geq \langle h_t, c \rangle_D - 2\epsilon \geq 1 - 2\Delta(A,C) - 4\alpha - 2\epsilon \ .$$

Therefore $\Delta(A, h) = (1 - \langle \phi, h \rangle_D)/2 \leq \Delta(A, C) + 2\alpha + \epsilon$.

Finally, we note that we assumed that $\gamma_i$ is known to the boosting algorithm. It is easy to see that we can use an appropriate estimate in the analysis above. Specifically, we use random samples to estimate the error of the weak hypothesis $g$ within $\gamma/4$. If the estimate is smaller than $1/2 - 3\gamma/4$ we update $h_i$ using $g$ with the empirical estimate in place of the true value $\gamma_i$. It is easy to see that in this case the distance will be reduced by at least $15\gamma^2/16$. The error estimate of $-\texttt{sign}(h_i)$ is treated analogously. □

We can now show that efficient weak agnostic learning with respect to distribution $D$ implies efficient agnostic learning with respect to distribution $D$.

**Theorem 3.2** *Let $C$ be a concept class and $D$ be a distribution over $X$ such that $C$ is efficiently weakly agnostically learnable over $D$. Then $C$ is efficiently agnostically learnable over $D$.*

**Proof:** By the definition, a weak agnostic learning algorithm gives a $(\tau, p(1/n, \tau))$-weak agnostic learning algorithm for every $\tau$ and some fixed polynomial $p(\cdot, \cdot)$. By boosting an $(\epsilon/3, p(1/n, \epsilon/3))$-weak agnostic learning algorithm using `A2boost` with the accuracy parameter set to $\epsilon/3$ we obtain an algorithm that outputs a hypothesis with performance $\Delta(A, C) + \epsilon$, in other words a strong agnostic learning algorithm. Note that the marginal distributions used in every stage of boosting are the same as in the original problem and the running time is polynomial in $n$ and $1/\epsilon$. □

An immediate application of Theorem 3.2 is a simple proof of agnostic learnability of decision trees over the uniform distribution and using membership queries that was recently obtained by Gopalan *et al.* [14].

**Lemma 3.3** *Let $C_s$ be the concept class of decision lists of size $s$ over $\{0,1\}^n$. $C_s$ is agnostically learnable over the uniform distribution and using membership queries in time polynomial in $n, s$ and $1/\epsilon$.*

**Proof:** As it has been shown by Kushilevitz and Mansour, the $L_1$ norm of the Fourier representation of a decision tree of size $s$ is at most $s$ [29]. Namely, if $c$ is a decision tree of size $s$ then $L_1(c) = \sum_{a \in \{0,1\}^n} |\hat{c}(a)| \leq s$, where $\hat{c}(a)$ is the Fourier coefficient of $c$ with index $a$. Now let $U$ be the uniform distribution over $\{0,1\}^n$, $\phi \in \mathcal{F}_1^\infty$ and let $A = (U, \phi)$. If $\Delta(A, C_s) \leq 1/2 - \tau$ then there exists $c \in C_s$ such that $\langle c, \phi \rangle_U \geq 2\tau$. But $c = \sum_{a \in \{0,1\}^n} \hat{c}(a) \chi_a(x)$ and, in particular,

$$\langle c, \phi \rangle_U = \sum_{a \in \{0,1\}^n} \hat{c}(a) \langle \chi_a(x), \phi \rangle_U \geq 2\tau \ .$$

This implies that there exists $a'$ such that $|\langle \chi_{a'}(x), \phi \rangle_U| \geq 2\tau/L_1(c) \geq 2\tau/s$. Therefore $\Delta(A, \chi_{a'}) \leq 1/2 - \tau/s$ or $\Delta(A, -\chi_{a'}) \leq 1/2 - \tau/s$. This implies that an agnostic learning algorithm for parity with $\epsilon = \tau/(2s)$ is also a weak agnostic learner for decision trees of size $s$. An agnostic learning algorithm for a parity function over the uniform distribution and using membership queries was given by Goldreich and Levin [13] (see also [29]). To finish the proof we simply need to apply Theorem 3.2. □

We now show that a simple modification to the distributions and the potential function used in `A2boost` gives an agnostic boosting algorithm from $(\alpha, \gamma)$-weak agnostic learning to $\alpha$-optimal agnostic learning.

**Theorem 3.4 (Restated from 1.1)** *There exists an algorithm `ABoost` that for every concept class $C$ and distribution $D$ over $X$, given an $(\alpha, \gamma)$-weak agnostic learning algorithm $\mathcal{A}$ for $C$ over $D$ agnostically and $\alpha$-optimally learns $C$ over $D$. Further, `ABoost` invokes $\mathcal{A}$ $O(\gamma^{-2})$ times and runs in time $\text{poly}(T, 1/\gamma, 1/\epsilon)$, where $T$ is the running time of $\mathcal{A}$.*

**Proof:** First we assume for simplicity that $\phi = f$ for some Boolean $f$, that is, the examples are labeled by a function. The proof is based on the same idea as the proof of Th. 3.1. However in order to avoid the double loss of $\alpha$ we use a different distribution $A_i$ at every stage. Specifically, we let $A_i = (D, P_1(f - h_i))$ while the rest of the algorithm is exactly the same. As before in order to prove the claim we first prove that the boosting process will terminate after at most $O(\gamma^{-2} + \epsilon^{-2})$ steps. The potential function whose gradient is $P_1(f - h_i)$ is defined as follows. For a real $a$ let

$$R(a) \triangleq \begin{cases} a^2 & |a| \leq 1 \\ 2|a| - 1 & \text{otherwise} \end{cases}$$

(It is easy to see that for every $b \in \{-1, 1\}$, $dR(b-a)/da = -P_1(b-a)$.) The potential of function $h \in \mathcal{F}_1^\infty$ relative to $f$ and $D$ is defined to be $\mathbf{E}_D[R(f-h)]$. We next claim that for every Boolean $f$ and distribution $D$,

1. $\mathbf{E}_D[R(f - h_0)] = \mathbf{E}_D[R(f)] = 1$ and for any real-valued function $\psi$, $\mathbf{E}_D[R(f - \psi)] \geq 0$;
2. If $\langle P_1(f - h_i), g_i \rangle_D \geq 2\gamma_i$ then $\mathbf{E}_D[R(f - (h_i + \gamma_i g_i))] \leq \mathbf{E}_D[R(f - h_i)] - 3\gamma_i^2$; to see this we simply observe that for every point $x$,

$$R(f(x) - (h_i(x) + \gamma_i g_i(x))) - R(f(x) - h_i(x)) \leq -2\gamma_i P_1(f(x) - h_i(x)) g_i(x) + (\gamma_i g_i(x))^2 \ .$$

3. $\mathbf{E}_D[R(f - P_1(h))] \leq \mathbf{E}_D[R(f - h)]$.

For the second part of the proof we prove that $\Delta(A, h) \leq \Delta(A, C) + \alpha + \epsilon/2$. As in the previous proof, $\langle f, c \rangle_D \geq 1 - 2\Delta(A, C)$. In addition, the stopping condition implies that $\langle P_1(f - h_t), c \rangle_D \leq 2\alpha$ and $\langle P_1(f - h_t), -\texttt{sign}(h_t) \rangle_D \leq \epsilon$. We now observe that $L_1$ norm of our distribution function $P_1(f - h_t)$ is small. Namely

$$\mathbf{E}_D[|P_1(f - h_t)|] = \mathbf{E}_D[f \cdot P_1(f - h_t)] = \mathbf{E}_D[c \cdot P_1(f - h_t)] + \mathbf{E}_D[(f - c) \cdot P_1(f - h_t)] \leq 2\alpha + \mathbf{E}_D[|f - c|]$$
$$\leq 2\alpha + 2\Delta(A, C) \ . \tag{1}$$

For the second step we show that

$$\Pr_D[f \neq \texttt{sign}(h_t)] = \frac{1}{2}\mathbf{E}_D[|f - \texttt{sign}(h_t)|] \leq \frac{1}{2}\mathbf{E}_D[P_1(f - h_t)(f - \texttt{sign}(h_t))] \leq \frac{1}{2}(\mathbf{E}_D[|P_1(f - h_t)|] + \epsilon). \tag{2}$$

By combining equations (1) and (2) we get that $\Pr_D[f \neq \texttt{sign}(h_t)] \leq \Delta(A, C) + \alpha + \epsilon/2$.

Finally, we note that if the function $\phi(x)$ is not Boolean we can reduce the analysis to the Boolean case by treating each point $x$ as two points: one with probability $D(x)(1 + \phi(x)/2)$ with the target function equal to 1 and the other one with probability $D(x)(1 - \phi(x)/2)$ and the target equal to $-1$. All functions that we consider are treated as identical on both of these points. The hypotheses we generate are combinations of the weak learning hypotheses and therefore will also be identical on both of these points. □

### 3.1 Relation to Distribution Independent Boosting

The boosting algorithms of this section can also be viewed in the regular setting where the boosting algorithm uses the weak learner on artificially constructed distributions over the domain. To see this one can observe that by outputting a random coin with expectation $(f(x) - h(x))/2$ (or $P_1(f(x) - h(x))$ in the case of ABoost) instead of $f(x)$ we reduce the contribution of the correlation on point $x$ to the total value of correlation in the same way as the regular boosting algorithms modify the weights of the point $x$ to reduce or increase the contribution. At the same time, as demonstrated by Ben-David *et al.* [2] and Gavinsky [12] modifying the distribution does give the boosting algorithm an ability to boost beyond an $\alpha$-optimal solution (at the expense of a stronger assumption). We demonstrate this by giving the following version of our boosting algorithm.

**Theorem 3.5** *There exists an algorithm* ABoostDI *that for every concept class $C$ over $X$, given a distribution independent $(\alpha, \gamma)$-weak agnostic learning algorithm $\mathcal{A}'$ for $C$, for every distribution $A = (D, f)$ over $X$ and $\epsilon > 0$, produces a hypothesis $h$ such that $\Pr_D[f \neq h] \leq \frac{\Delta(A,C)}{1-2\alpha} + \epsilon$. Further,* ABoost *invokes $\mathcal{A}'$ $O(\gamma^{-2}\Delta'^{-1}\log(1/\Delta'))$ times for $\Delta' = \Delta(A, C)/(1 - 2\alpha)$ and runs in time $\text{poly}(T, 1/\gamma, 1/\epsilon)$, where $T$ is the running time of $\mathcal{A}'$.*

**Proof:** We first observe that for every Boolean function $f, h \in \mathcal{F}_1^\infty$, and $x \in X$, $P_1(f(x) - h(x)) = f(x)|P_1(f(x) - h(x))|$. This implies that for every function $g \in \mathcal{F}_1^\infty$,

$$\mathbf{E}_D[P_1(f(x) - h(x))g(x)] = \mathbf{E}_D[|P_1(f(x) - h(x))|f(x)g(x)] = \mathbf{E}_{D_h}[f(x)g(x)] \cdot N_h ,$$

where $D_h$ is the distribution defined by the density function $D_h(x) = D(x)|P_1(f(x) - h(x))|/N_h$ and $N_h = \mathbf{E}_D[|P_1(f(x) - h(x))|]$ is the normalization factor. Therefore if $\mathcal{A}'$ provides a hypothesis that satisfies $\mathbf{E}_{D_h}[f(x)g(x)] \geq 2\gamma$ then $\mathbf{E}_D[P_1(f(x) - h(x))g(x)] \geq 2\gamma \cdot N_h$. In order to bound the number of boosting stages we need to lower bound $\gamma \cdot N_h$. To do this we note that $\mathbf{E}_D[|P_1(f - h)|] \geq \Pr[f \neq \text{sign}(h)]$ and therefore we can assume that $N_h \geq \Delta'$ which is the desired final error of the boosting algorithm that we will determine later. Otherwise the error of $\text{sign}(h)$ is already sufficiently and we can stop the boosting (using an additional testing step). This implies that the total number of calls to the weak learner is $O(\gamma^{-2}\Delta'^{-2})$. We can also sharpen this bound by noticing that $N_h = \mathbf{E}_D[|P_1(f-h)|] \geq E_D[R(f-h)]/3$. This implies that at every stage when the weak learner's hypothesis is used

$$\mathbf{E}_D[R(f - h_{i+1})] \leq \mathbf{E}_D[R(f - h_i)] - 3(\gamma \cdot N_h)^2 \leq \mathbf{E}_D[R(f - h_i)](1 - \gamma^2 \cdot N_h) \leq \mathbf{E}_D[R(f - h_i)](1 - \gamma^2 \cdot \Delta') .$$

This implies that after at most $t = \gamma^{-2}\Delta'^{-1} \ln(\Delta'^{-1})$ steps $\Pr[f \neq \text{sign}(h_t)] \leq \mathbf{E}_D[R(f - h_t)] \leq \Delta'$, which implies that the boosting process will terminate.

Finally, we need to define $\Delta'$. By the definition, $\mathcal{A}'$ returns a good weak learner whenever $\mathbf{E}_{D_h}[f(x)g(x)] \geq 2\alpha$ and hence if the boosting stopped then $\mathbf{E}_D[c \cdot P_1(f - h_t)] < 2\alpha \cdot N_{h_t}$. By plugging this into equation (1), we obtain that

$$N_{h_t} = \mathbf{E}_D[|P_1(f - h_t)|] \leq 2\Delta(A, C) + 2\alpha \cdot N_{h_t},$$

which gives us the bound $N_{h_t} \leq \frac{2\Delta(A,C)}{1-2\alpha}$. By plugging this into equation (2) we obtain that

$$\Pr_D[f \neq \text{sign}(h_t)] \leq N_{h_t}/2 + \epsilon/2 \leq \frac{\Delta(A,C)}{1-2\alpha} + \epsilon/2 .$$

We therefore set $\Delta' = \frac{\Delta(A,C)}{1-2\alpha} + \epsilon/2$. $\square$

To compare this result with the boosting algorithms in the setting of Ben-David *et al.* [2] we note that, by the definition, any $\beta$-optimal agnostic learner is in particular, a $(\beta + \gamma', \gamma'/2)$-weak agnostic learner for any inverse-polynomial $\gamma' \geq 0$. Therefore ABoostDI applied to a $\beta$-optimal agnostic learner returns a hypothesis with the error of at most $\frac{\Delta(A,C)}{1-2\beta+\gamma'} + \epsilon/2 \leq \frac{\Delta(A,C)}{1-2\beta} + \epsilon$ for $\gamma' = \epsilon/(4(1-2\beta))$. As demonstrated by Gavinsky, this is optimal [12].

**Remark 3.6** *We remark that* ABoostDI *does not modify the target function and therefore is also applicable in the PAC framework. In addition,* ABoostDI *has the optimal* smoothness. *That is, when learning to accuracy $\epsilon'$ the weight of any point under any of the distribution generated by the boosting algorithm is at most $1/(2\epsilon' - \epsilon)$ times higher than the weight under $D$. This is true since $D_h(x) = D(x)|P_1(f(x) - h(x))|/N_h$ and $N_h \geq 2\Pr[f \neq \text{sign}(h)] - \epsilon$. Smoothness property is crucial in a number of applications of boosting algorithms such as learning DNF over the uniform distribution [19], learning with malicious noise [32] and the connection to hard-core set constructions [27].*

## 3.2 Applications to PAC Learning

It has long been noted that efficient agnostic learning of a concept class $C$ over a distribution $D$ implies efficient *weak* learning of the class of functions expressible as low-weight linear thresholds of functions from $C$ over $D$ [25]. It therefore follows that distribution-independent agnostic learning of $C$ implies PAC learning of $\text{TH}(W, C)$ (see Section 1.1 for the definition) for any polynomially bounded $W$. Our goal is to strengthen this implication to distribution-specific learning.

**Theorem 3.7 (Restated from 1.2)** *If $C$ is agnostically learnable with respect to distribution $D$ in time $T(n, \epsilon)$ then $\text{TH}(W, C)$ is PAC learnable over $D$ in time $O((W/\epsilon)^2 \cdot T(n, \epsilon/(4W)) + poly(n, W, 1/\epsilon))$.*

**Proof:** The reduction from PAC to agnostic learning relies on the discriminator lemma of Hajnal *et al.* [15] stating that for every $f \in \text{TH}(W, C)$ and every distribution $D$, there exists a function in $c' \in C$ such that $|\langle f, c' \rangle_D| \geq 1/W$. Now for $h \in \mathcal{F}_1^\infty$, let $D_h$ be the distribution defined in the proof of Th. 3.5 (that is $D_h(x) = D(x)|P_1(f(x) - h(x))|/N_h$). The discriminator lemma implies that there exists $c'$ such that $|\langle f, c' \rangle_{D_h}| \geq 1/W$. This implies that $|\langle f - h, c' \rangle_D| \geq N_h/W$. As we have showed in Th. 3.5, $N_h \geq \Pr_D[f \neq h] \geq \epsilon$ since the desired accuracy of PAC learning is $\epsilon$. This implies that $|\langle f - h, c' \rangle_D| \geq \epsilon/W$. Therefore when the agnostic learner for $C$ is run on the distribution $(D, f - h)$ with $\epsilon' = \epsilon/(4W)$ it will return a function $g$ such that $\langle f - h, g \rangle_D \geq \epsilon/(2W)$ (for simplicity we can assume that $C$ is closed under negation; alternatively we can also run the agnostic learner on the negation of $(D, f - h)$ and negate the result). We can therefore use $g$ in the same way as in the Theorem 1.1 with $\gamma = \epsilon/(4W)$ and the accuracy of the boosting process set to $\epsilon'' = \epsilon/2$ until the accuracy of $\text{sign}(h_i)$ reaches $\epsilon$. The total number of boosting stages is $O((W/\epsilon)^2)$ and the running time is polynomial in $1/\epsilon$ and $W$ and the running time of the agnostic learner for $C$ (with $\epsilon' = \epsilon/(4W)$). Finally we note that in this result one can also use the slightly simpler `A2boost` in place of `ABoost` and there is not need to balance $h_i$ by testing $-\text{sign}(h_i)$ hypothesis (since those only affect the constant multiplier of the accuracy). □

One application of such a result is that it gives a DNF learning algorithm directly from a uniform distribution agnostic parity learning algorithm (such as the Kushilevitz-Mansour algorithm [29]) without the need for a specialized analysis of the Fourier Transform of distribution functions given by Jackson [19] and used in many subsequent works. This follows from the fact that polynomial size DNF formulas can be represented as low-weight thresholds of parities (or TOP) [19]. We further note that the resulting algorithm is the same as that obtained in Lemma 3.3 (up to the setting of the parameters).

A related corollary of this result is that agnostic learning of DNF formulas from membership queries over the uniform distribution (see [14] for the problem definition) would imply (strong) learning of depth-3 circuits (even with a majority gate at the top) with membership queries.

## 4 Relation to Hard-core Set Construction

We start with a couple of definitions relevant to hard-core set construction. We say that a function $f$ is $\lambda$-hard for size $s$ if for every circuit $z$ of size at most $s$, $\Pr_U[f(x) \neq z(x)] \geq \lambda$, where $U$ denotes the uniform distribution over $X$. A *measure* $M$ over $X$ is a function from $X$ to $[0, 1]$. The density of a measure $M$ is defined to be $\mu(M) = (\sum_{x \in X} M(x))/|X|$. We say that $f$ is $\gamma$-hard-core on $M$ for size $s$ if for every circuit $z$ of size at most $s$, $\Pr_{U_M}[f(x) \neq z(x)] \geq 1/2 - \gamma$, where $U_M$ is the distribution on $X$ with density function $U_M(x) = M(x)/\mu(M)$. Similarly, we say that $f$ is $\gamma$-hard-core on a set $S \subseteq X$ for size $s$ if $f$ is $\gamma$-hard-core on $M_S$ for size $s$, where $M_S(x)$ is the characteristic function of $S$. It is well-known that in order to construct a hard-core set of size $\delta \cdot |X|$ it is sufficient to construct a hard-core measure of density at least $\delta$ [18]. All known *uniform* constructions of a measure for which $f$ is $\gamma$-hard-core are essentially boosting algorithms that construct a sequence of measures $M_0, M_1, \ldots$ each of density at least $\delta$ such that if $f$ is not $\gamma$-hard-core on $M_i$ for size $s$ then a circuit $z_i$ that $(1/2 - \gamma)$-approximates $f$ on $M_i$ is used to create $M_{i+1}$. If this process does not stop after a certain number of steps then $z_i$'s can be combined to obtain a circuit that $\lambda$-approximates $f$. This contradicts $\lambda$-hardness of $f$ for some size $s'$ and therefore implies that $f$ is $\gamma$-hard-core on one of the constructed measures. An important parameter of such a construction is the density $\delta$ as a function of $\lambda$ (and sometimes $\gamma$). Impagliazzo showed a construction of a hard-core set of size $\lambda$ and asked whether the optimal size of $2\lambda$ is achievable [18]. Holenstein [17] gave the first construction with the optimal hard-core set size parameter on the basis of Impagliazzo's hard-core set construction [18]. In a recent work Barak *et al.* gave a more efficient construction based on multiplicative updates and Bregman projections [1].

It is easy to see that the distribution $U_M$ is $1/\delta$-smooth if and only if the measure $M$ has density $\delta$ [27]. Gavinsky has demonstrated that the smoothness of distributions produced by a boosting algorithm determines the error that the boosting algorithm will achieve when boosting a (distribution-independent) $\beta$-optimal agnostic learner [12]. Our goal is to combine these observations in the context of distribution-specific agnostic boosting. Namely we are going to show that hard-core set constructions that achieve the optimal set size of $2\lambda$ are (distribution-specific) agnostic boosting algorithms. We start with a formal statement of the hard-core set lemma with the optimal set size parameter.

**Lemma 4.1 ([17, 1])** *Let $f$ be a Boolean function over a domain $X$, $s$ be an integer, $\delta > 0$ and $\gamma > 0$. Suppose, there exists an algorithm $\mathcal{A}$ which for any measure $M$ over $X$ of density $\delta$, given access to random examples from distribution $(U_M, f)$, returns a circuit $z$ of size at most $s$ such that $\Pr_{U_M}[z(x) \neq f(x)] \leq 1/2 - \gamma$. Then there is an algorithm $\mathcal{B}$ which for every $f$, given access to random examples from distribution $(U, f)$, with probability at least $1/2$ (over the internal randomness of $\mathcal{B}$) returns a circuit $z'$ of size $s'$ such that $\Pr_U[z'(x) \neq f(x)] \leq \delta/2$. Furthermore, the algorithm $\mathcal{B}$ invokes $\mathcal{A}$ $t(1/\delta, 1/\gamma)$ times; requires time $m(1/\delta, 1/\gamma, s, T)$ and $s'$ is linear in $t(1/\delta, 1/\gamma)$ and $s$, where $T$ is the time to simulate $\mathcal{A}$, and $t(\cdot, \cdot)$ and $m(\cdot, \cdot)$ are fixed polynomials.*

First, while this lemma is stated for the uniform distribution on $X$ it is known and easy to verify directly that it also holds for any distribution $D$ over $X$ (implicit in [1] and in general can be obtained by taking a sufficiently large sample from $D$ and viewing a uniform distribution over the sample). Namely, the lemma holds even if one replaces the uniform distribution by $D$, density by density relative to $D$, that is $\mu_D(M) = \mathbf{E}_D[M(x)]$, and $U_M$ with the distribution $D_M$ defined by density function $D_M(x) = D(x)M(x)/\mu_D(M)$.

Now let $C$ be a concept class, $A = (D, f)$ be a distribution over examples and $\mathcal{A}'$ be an $(\alpha, \gamma)$-weak agnostic learner $\mathcal{A}'$ for $C$ over $D$. To obtain an agnostic boosting algorithm we replace the algorithm $\mathcal{B}$ in the lemma with $\mathcal{A}'$. To do this we generate examples from distribution $A_M = (D, f \cdot M)$ and run $\mathcal{A}'$ on them. We then return the hypothesis $g$ given by $\mathcal{A}'$ to the algorithm $\mathcal{A}$. We claim that if $\mu_D(M) \geq 2(\Delta(A, C) + \alpha)$ then $\Pr_{D_M}[g(x) \neq f(x)] \leq 1/2 - \gamma$. Observe that if the claim holds then the execution of $\mathcal{A}$ will produce a hypothesis $z'$ such that $\Pr_D[z'(x) \neq f(x)] \leq \Delta(A, C) + \alpha$ as desired. At the same time the running time is polynomial in the relevant parameters of agnostic learning (in fact it does not even depend on the accuracy $\epsilon$).

To establish the claim we let $c$ be the concept such that $\Delta(A, C) = \Delta(A, c) = \Pr_D[f \neq c]$. Now

$$\mathbf{E}_D[c \cdot (f \cdot M)] = \mathbf{E}_D[f \cdot (f \cdot M)] + \mathbf{E}_D[(c - f) \cdot (f \cdot M)] \leq \mathbf{E}_D[M] - 2\Pr_D[|c - f|] = \mu_D(M) - 2 \cdot \Delta(A, C) \ .$$

Therefore if $\mu_D(M) \geq 2(\Delta(A, C) + \alpha)$ then $\mathbf{E}_D[c \cdot (f \cdot M)] \geq 2\alpha$ or $\Delta(A_M, c) \leq 1/2 - \alpha$. Hence, by the definition of $\mathcal{A}'$, it will return a hypothesis $g$ such that $\mathbf{E}_D[g \cdot (f \cdot M)] \geq 2\gamma$. This gives us that

$$\mathbf{E}_{D_M}[g \cdot f] = \mathbf{E}_D[g \cdot (f \cdot M)]/\mu_D(M) \geq 2\gamma/\mu_D(M) \geq 2\gamma \ ,$$

or $\Pr_{D_M}[g(x) \neq f(x)] \leq 1/2 - \gamma$.

Finally, we also observe that `ABoostDI` gives a hard-core set construction that achieves essentially optimal hard-core set size. As we have showed in Remark 3.6, when learning to accuracy $\epsilon'$ the algorithm `ABoostDI` produces $1/(2\epsilon' - \epsilon)$-smooth distributions for any $\epsilon$ (in time polynomial in $1/\epsilon$). By the observation of Klivans and Servedio [27], this implies that when used for hard-core set constructions the algorithm will produce a set of size $2\epsilon' - \epsilon$ distributions for any $\epsilon$ in time polynomial in $1/\epsilon$.

## 5 Conclusions

We demonstrated that in the agnostic learning framework strong learning with respect to a specific distribution can be efficiently reduced to weak learning with respect to the same distribution. Further, we showed that this can be done using a variety of methods, some new and simple ones given here but also via a simple adaptation of two known algorithms [17, 1] (and yet another method was just discovered independently [20]). In our opinion these findings testify that boosting in the agnostic learning framework is at least as natural and powerful phenomenon as it is the PAC model. The agnostic learning model reflects many of the practical scenarios more faithfully than the PAC model [16] and hence we suggest that the agnostic learning framework is better suited for theoretical analysis of boosting algorithms.

One evidence for this is that the main reason why the Adaboost algorithm [11] is known not to cope well with noise is that it places too much weight on the noisy examples [6], in other words it is not smooth. As can be seen from our work (and from [12]), achieving the strong agnostic guarantees forces the boosting algorithm to be optimally smooth.

In this version of the results we have omitted the discussion of the circuits that combine the weak hypotheses and also detailed bounds on the running time of our boosting algorithms. In part, this is because our algorithms do not improve on the agnostic boosting algorithm derived from the algorithm of Barak *et al.* [1] that achieves the optimal number of boosting stages and uses a simple majority to combine the weak hypotheses. While the performance of `ABoost` in the distribution-specific setting is essentially the same, our algorithm uses a more complex circuit to combine the weak hypotheses (primarily because of the "balancing" step). In addition, this allows us to simplify the presentation of the core ideas of the algorithm.